\documentclass[10pt, conference, compsocconf]{IEEEtran}  

\usepackage{amsmath} 
\usepackage{bm}
\usepackage{amssymb}
\usepackage{siunitx}
\usepackage[pdftex]{graphicx}
\usepackage{epstopdf}
\usepackage{float}
\usepackage{cite}
\usepackage{color}
\usepackage{cancel}
\usepackage{dblfloatfix}
\usepackage{soul}
\usepackage{multirow}
\usepackage{subcaption}

\IEEEoverridecommandlockouts                              
\usepackage[T1]{fontenc}
\newcommand{\norm}[1]{\ensuremath{\left\Vert#1\right\Vert}}
\newcommand{\f}{\frac}
\DeclareMathAlphabet{\mbf}{OT1}{ptm}{b}{n}
\newcommand{\trans}{{\ensuremath{\mathsf{T}}}}
\newcommand{\kmin}{{k-1}}
\newcommand{\mbs}[1]{\ensuremath{\boldsymbol{#1}}}
\newcommand{\mbfdot}[1]{\ensuremath{\dot{\mbf{#1}}}}

\newcommand{\mbfbar}[1]{\ensuremath{\bar{\mbf{#1}}}}
\newcommand{\mbfhat}[1]{\ensuremath{\hat{\mbf{#1}}}}
\newcommand{\mbfcheck}[1]{\ensuremath{\check{\mbf{#1}}}}

\newcommand{\mbftilde}[1]{\ensuremath{\tilde{\mbf{#1}}}}
\newcommand{\eye}{\mbf{1}}
\newcommand{\zero}{\mbf{0}}

\newcommand{\spliteqindent}{&\qquad\qquad\qquad}

\newcommand{\diag}{{\ensuremath{\mathrm{diag}}}}

\DeclareMathOperator{\Exp}{Exp}
\DeclareMathOperator{\Log}{Log}

\title{\LARGE \textbf{Towards Open World NeRF-Based SLAM}
}

\author{\IEEEauthorblockN{Daniil Lisus$^{*}$, Connor Holmes$^{*}$, and Steven Waslander}
\IEEEauthorblockA{University of Toronto Robotics Institute\\
University of Toronto\\
Toronto, Canada\\
Email: [FIRSTNAME].[LASTNAME]@robotics.utias.utoronto.ca}
\thanks{$^*$Authors contributed equally to this work.}
}

\begin{document}

\newpage
%
%
%
%
%
%
%
\def \myJournal {2023 20th Conference on Robots and Vision (CRV)}
\def \myDoi {10.1109/CRV60082.2023.00013}
\def \myPaperSiteName {IEEE Xplore}
\def \myPaperSiteLink {https://ieeexplore.ieee.org/document/10229827}
\def \myYear {2023}

\def \myPaperCitation{D. Lisus, C. Holmes and S. Waslander, ``Towards Open World NeRF-Based SLAM,'' \textit{2023 20th Conference on Robots and Vision (CRV)}, Montreal, QC, Canada, 2023, pp. 37-44.}

\begin{figure*}[t]

\thispagestyle{empty}
\begin{center}
\begin{minipage}{6in}
\centering
This paper has been accepted for publication in \emph{\myJournal}. 
\vspace{1em}

This is the author's version of an article that has, or will be, published in this journal or conference. Changes were, or will be, made to this version by the publisher prior to publication.
\vspace{2em}

\begin{tabular}{rl}
DOI: & \myDoi\\
\myPaperSiteName: & \texttt{\myPaperSiteLink}
\end{tabular}

\vspace{2em}
Please cite this paper as:

\myPaperCitation

\vspace{15cm}
\copyright \myYear \hspace{4pt}IEEE. Personal use of this material is permitted. Permission from IEEE must be obtained for all other uses, in any current or future media, including reprinting/republishing this material for advertising or promotional purposes, creating new collective works, for resale or redistribution to servers or lists, or reuse of any copyrighted component of this work in other works.

\end{minipage}
\end{center}
\end{figure*}
\newpage
\clearpage
\pagenumbering{arabic} 

\maketitle
\thispagestyle{empty}
\pagestyle{empty}

\begin{abstract}
Neural Radiance Fields (NeRFs) offer versatility and robustness in map representations for Simultaneous Localization and Mapping (SLAM) tasks. This paper extends NICE-SLAM, a recent state-of-the-art NeRF-based SLAM algorithm capable of producing high quality NeRF maps. However, depending on the hardware used, the required number of iterations to produce these maps often makes NICE-SLAM run at less than real time. Additionally, the estimated trajectories fail to be competitive with classical SLAM approaches. Finally, NICE-SLAM requires a grid covering the considered environment to be defined prior to runtime, making it difficult to extend into previously unseen scenes. This paper seeks to make NICE-SLAM more open-world-capable by improving the robustness and tracking accuracy, and generalizing the map representation to handle unconstrained environments. This is done by improving measurement uncertainty handling, incorporating motion  information, and modelling the map as having an explicit foreground and background. It is shown that these changes are able to improve tracking accuracy by 85\% to 97\% depending on the available resources, while also improving mapping in environments with visual information extending outside of the predefined grid.
\end{abstract}


\section{Introduction}
Starting with the landmark paper by Mildenhall et al. \cite{mildenhall_nerf_2020}, Neural Radiance Fields (NeRFs) have taken the machine vision and robotics perception communities by storm. The central idea behind NeRF is to combine classical graphics rendering techniques with a multilayer perceptron (MLP) trained on image data to learn an \emph{implicit} representation of a given scene. The scene can then be rendered from novel viewpoints (i.e., view synthesis). Within the context of robotics, this approach holds promise to address shortcomings of classical dense SLAM algorithms. In particular, NeRFs are well suited to estimate portions of the map for unobserved regions \cite{zhu_nice-slam_2022}. They can also be leveraged to view maps from perspectives that may be interesting to users, but which have not been directly visited by the robot. Since they are fundamentally based on MLPs, NeRF maps can also be trained to be robust to changes in map environment conditions such as lighting or time of the year \cite{rudnev_nerf_2022}.

The original formulation of NeRF, which used one large MLP, required hours of training, was slow to render, and required exact knowledge of input camera poses. However, a number of advancements since have considerably improved on all of these issues. Several authors have shown that encoding the NeRF in a spatial data structure leads to considerable improvements in speed and accuracy, often using smaller MLPs to decode spatial features during view synthesis \cite{takikawa_neural_2021,yu_plenoxels_2021,muller_instant_2022}. In particular, NGLOD \cite{takikawa_neural_2021} proposed the use of small MLPs in a volumetric grid, while Plenoxels \cite{yu_plenoxels_2021} used a spatial octree and bypassed MLP use altogether. These ideas were further improved upon by using spatial hash tables to attain NeRFs that can be trained in real-time \cite{muller_instant_2022}. BARF \cite{lin_barf_2021} and NeRF-\,- \cite{wang_nerf--_2022} have also shown that a priori exact pose knowledge is unnecessary to reproduce both accurate pose estimates and NeRFs. 

These advancements have opened the door for the use of NeRFs to represent maps for Simultaneous Localization and Mapping (SLAM) in robotics applications. One of the first papers in this area was iMAP \cite{sucar_imap_2021}, which built an MLP-based NeRF in real-time using RGB-D data by downsampling the number of pixels used to generate the NeRF. NICE-SLAM \cite{zhu_nice-slam_2022} improved significantly on this framework by leveraging the key insight that the \textit{entire} NeRF map need not be updated at every iteration. By introducing a spatial, voxel-based NeRF, NICE-SLAM only updates parts of the NeRF that are spatially relevant to a given camera view. Vox-Fusion was a similar, concurrent work to NICE-SLAM, but with a dynamically allocated voxel grid. Released after the start of this work, NeRF-SLAM \cite{rosinol_nerf-slam_2022} has further developed the concepts presented in NICE-SLAM and extended them to monocular SLAM.

Although NICE-SLAM produces complete, dense maps, it fails to perform competitively in the generated pose estimates compared to classical SLAM approaches. Additionally, the produced volumetric grid can become very expensive to maintain if the algorithm aims to function in a large environment. Looking forward to potential open world deployment of NeRF-based SLAM, all current approaches make use of a predefined, finite volumetric grid, without a clear way to handle visual information contained far from the camera or to dynamically expand the area of operation.

This paper begins to tackle these problems in order to leverage the special mapping approach of NeRF-based SLAM without sacrificing the quality of the trajectory estimates or requiring excessive computational resources. Specifically, the contributions of this paper are
\begin{itemize}
    \item including depth uncertainty from RGB-D images in all depth loss terms to improve local accuracy,
    \item implementing motion information to improve camera tracking and handle difficult motion, and
    \item separating the NeRF into a finite foreground grid and a background sphere to handle environments of any size.
\end{itemize}
Although this paper deals specifically with RGB-D information, these extensions could also be applied in other NeRF-based SLAM methods (e.g., monocular methods \cite{rosinol_nerf-slam_2022}). 

\section{Background}
\subsection{NICE-SLAM Algorithm}
This section provides a brief review of the NICE-SLAM algorithm. Similar to its predecessor, iMAP, and other modern SLAM algorithms \cite{orb_slam}, NICE-SLAM separates the phases of SLAM into two parallel threads: a \textit{tracking} thread to localize the current camera frame against the current map and a \textit{mapping} thread that jointly optimizes the parameters of the NeRF map and a set of stored \textit{keyframes}.

NICE-SLAM uses a series of 3 fixed-size, voxel grids of encoded features with different grid resolutions to represent the map. When evaluating pixels for a given RGB-D image, NICE-SLAM uses the same ray-casting technique as \cite{mildenhall_nerf_2020}, but evaluates the rays by interpolating sample points in the voxel grid (similar to \cite{yu_plenoxels_2021}), decoding each sample feature with a (smaller) MLP and aggregating the result into an estimated pixel color and depth. Additional details on mapping and tracking threads are given in the subsections below.


\subsubsection{Mapping}
The mapping thread is responsible for updating the voxel-grid NeRF representation of the map. It does this by continually optimizing the voxel-grid features and the stored poses of a relevant subset of \textit{keyframes}. Similar to ORB-SLAM \cite{orb_slam}, keyframes are selected on the basis of the level of information gain that they provide and consist of an RGB-D image as well as the corresponding estimated camera pose. 
For each map update, a set of keyframes is selected based on predicted \textit{overlap} with the current frame and are used, along with the current frame, to construct a loss function. 

The loss function has both depth-based and colour-based components. The depth-based loss for each grid resolution in the mapping thread based on $N$ pixels is given by
\begin{align}
    \mathcal{L}_\mathrm{depth}^\mathrm{map} = \sum_{r=f,c}\f{1}{N} \sum_{n=1}^N \norm{d_n - \hat{d}_{r,n}}_1, \label{eq:mapping_depth}
\end{align}
where $d_n$ is the measured pixel depth and $\hat{d}_{r,n}$ is the NeRF-predicted pixel depth for the fine ($f$) and coarse ($c$) grid resolutions. The color-based loss is only generated at the fine grid resolution and is given by
\begin{align}
    \mathcal{L}_\mathrm{colour}^\mathrm{map} = \f{1}{N} \sum_{n=1}^N \norm{I_n - \hat{I}_n}_1, \label{eq:mapping_color}
\end{align}
where $I_n$ is the measured pixel colour and $\hat{I}_n$ is the NeRF-predicted pixel colour. The overall optimization objective for the mapping portion is given by
\begin{equation}
    \begin{array}{rl}
        \min\limits_{\mbf{C}_i,\mbf{r}_i,\mbs{\Theta}} & \sum_{i=1}^{M} \mathcal{L}_\mathrm{depth}^\mathrm{map,i} + \lambda_m \mathcal{L}_\mathrm{colour}^\mathrm{map,i},
    \end{array}
\end{equation}
where $ i $ represents the $i^{th}$ of $M$ keyframes, $\mbf{C}$ and $\mbf{r}$ are the camera orientation and position respectively, $\mbs{\Theta}$ are the voxel grid features, and $\lambda_m$ is a tuning parameter.

\subsubsection{Tracking}
The tracking thread performs a similar optimization to the mapping thread, but only optimizes the pose of the \textit{current} frame (i.e. does not modify the map).

The colour component of the loss, $\mathcal{L}_\mathrm{colour}^\mathrm{track}$, is computed in the same way as \eqref{eq:mapping_color}, but only considers the current frame. On the other hand, the depth-based loss in the tracking thread is weighted based on depth uncertainty in the network. For each grid resolution, the loss is computed for  $N$ pixels as
\begin{align}
\mathcal{L}_\mathrm{depth}^\mathrm{track} = \sum_{r=f,c}\f{1}{N} \sum_{n=1}^N \f{\norm{d_n - \hat{d}_{r,n}}_1}{\hat{\sigma}^d_{r,n}}, \label{eq:tracking_depth}
\end{align}
where variables are defined in the same way as for \eqref{eq:mapping_depth}, with $\hat{\sigma}^d_n$ corresponding to the standard deviation on $\hat{d}_n$. This standard deviation is extracted from the NeRF voxel grid by computing the variance from all points that contribute to the final $\hat{d}_n$ value along a pixel ray. The overall tracking optimization objective is given by
\begin{equation}
    \begin{array}{rl}
        \min\limits_{\mbf{C},\mbf{r}} & \mathcal{L}_\mathrm{depth}^\mathrm{track} + \lambda_t \mathcal{L}_\mathrm{colour}^\mathrm{track},
    \end{array}
\end{equation}
where $\mbf{C}$ and $\mbf{r}$ are the pose parameters for the current frame and $\lambda_t$ is a tuning parameter.

\subsection{$SO(3)$ Preliminaries}
The camera orientation in this paper is represented by the $SO(3)$ matrix Lie group (MLG). Operations related to $SO(3)$ are presented below, but the full MLG theory is omitted for brevity. Please refer to \cite{microLieTheory} for more information. The logarithmic ($\log(\cdot)$) and exponential ($\exp(\cdot)$) operators convert an $SO(3)$ element to and from the Lie algebra. The vee ($(\cdot)^\vee$) and wedge ($(\cdot)^\wedge$) operators convert elements of the Lie algebra to and from the $\mathbb{R}^3$ representation, for the $3$ degrees of freedom of $SO(3)$. The capital logarithmic and exponential operators for an $SO(3)$ element $\mbf{C}$ and its $\mathbb{R}^3$ representation $\mbs{\xi}$ are defined as
\begin{align}
    \mbf{C} &= \exp(\mbs{\xi}^\wedge) \triangleq \Exp(\mbs{\xi}) \in SO(3),\\
    \mbs{\xi} &= \log(\mbf{C})^\vee \triangleq \Log(\mbf{C}) \in \mathbb{R}^3.
\end{align}
Additionally, the right-invariant error definition for $SO(3)$ is adopted here arbitrarily. With this definition, the difference between a nominal $\mbfbar{C}$ and $\mbf{C}$ is computed as $\delta\mbf{C} = \mbfbar{C}\mbf{C}^\trans$.

\section{Open World Improvements}
The baseline NICE-SLAM algorithm is modified to bring it closer to being able to handle open world environments.

\subsection{Depth Uncertainty}
The standard NICE-SLAM algorithm does not consider depth uncertainty from the measured RGB-D depth $d_n$. As a result, when considering $N$ pixels, each measured depth pixel contributes equally to the computed depth loss in \eqref{eq:mapping_depth} and \eqref{eq:tracking_depth}. While \eqref{eq:tracking_depth} does down-weight the contributions according to a proxy measure of the uncertainty on the NeRF-generated depth value, it still treats the RGB-D depth values as equally valuable. This equal valuation is contrary to not only potential sensor-specific uncertainty fluctuations or biases, for example due to artifacts such as vignetting, but also to the fact that vision-based depth sensors produce measurements that typically increase in uncertainty as a function of the depth. Although often neglected in relatively constant depth environments, this uncertainty can become considerable in larger scenes. To account for this varying uncertainty, the depth losses in \eqref{eq:mapping_depth} and \eqref{eq:tracking_depth} are modified as
\begin{align}
    \mathcal{L}_\mathrm{depth}^\mathrm{map} &= \f{1}{N} \sum_{n=1}^N \f{\norm{d_n - \hat{d}_n}_1}{{\sigma}_n^d}, \label{eq:new_tracking_depth},\\
    \mathcal{L}_\mathrm{depth}^\mathrm{track} &= \f{1}{N} \sum_{n=1}^N \f{\norm{d_n - \hat{d}_n}_1}{\sqrt{{{\sigma}^{d^2}_n} + {\hat{\sigma}_n^{d^2}}}}, \label{eq:new_mapping_depth}
\end{align}
where ${\sigma}_n^d$ is the standard deviation of $d_n$, the depth measured in pixel $n$. For simplicity, the uncertainties from $d_n$ and $\hat{d}_n$ are combined assuming the two variables are uncorrelated. Finding the correlated uncertainty in this term requires a probabilistic NeRF map and is left for future work.

\subsection{Including Motion Data}
Motion data can serve as a valuable source of information for state estimation. This paper includes motion data in both the tracking and mapping optimization to make use of the temporal relationship between camera poses. This data is typically provided by an inertial measurement unit (IMU). However, datasets that have thus far been used to evaluate the performance of NeRF-based SLAM methods have not contained IMU measurements. As a result, motion data needs to be generated from ground truth poses. Since generating acceleration in this manner incurs a lot of error, owing to the double difference, the considered ``IMU'' measurements are linear velocity $\mbf{v}$ and angular velocity $\mbf{u}$. It is assumed that these measurements are subject to additive Gaussian noise $\mbf{w}^\mbf{m} \sim \mathcal{N}(\mbf{0}, \mbs{\Sigma}^\mbf{m})$ for each measurement $\mbf{m} \in [\mbf{v}, \mbf{u}]$. The approach presented here can be extended to linear acceleration in a similar manner.

\subsubsection{Tracking with Motion Data}
An IMU-based loss is added to the tracking objective that provides a motion prior for the current camera pose based on its immediate predecessor. As a simplification, the previous camera pose is treated as fixed, with the IMU-based loss being weighted by a factor corresponding to the IMU noise. The camera pose at time $t_k$ is defined as $\mbf{X}_k = (\mbf{C}_k, \mbf{r}_k)$, where $\mbf{C} \in SO(3)$ is the camera orientation and $\mbf{r} \in \mathbb{R}^3$ is the position.

Consider the current camera pose $\mbf{X}_k$ at time $t_{k}$ and the previous camera pose $\mbf{X}_\kmin$ at time $t_\kmin$. The IMU loss is composed of an orientation and position component. The orientation component error is
\begin{align}
    \mbf{e}_\mathrm{IMU}^{\mbf{C}} = \Log\left(\Exp(\mbf{u}_\kmin T_\kmin)\mbf{C}_k^\trans\mbf{C}_\kmin\right).
\end{align}
For the final loss, this error is weighted by the uncertainty from $\mbf{u}_\kmin$. This weight is computed by linearizing $\mbf{e}_\mathrm{IMU}^{\mbf{C}}$ with respect to the measurement noise $\mbf{w}^\mbf{u}_\kmin$ and propagating the noise covariance $\mbs{\Sigma}_\kmin^\mbf{u}$ through the linearization as
\begin{align}
    \mbs{\Sigma}_\mathrm{IMU}^\mbf{C} = \left(\f{\partial\mbf{e}_\mathrm{IMU}^{\mbf{C}}}{\partial\mbf{w}^\mbf{u}_\kmin}\right) \mbs{\Sigma}_\kmin^\mbf{u} \left(\f{\partial\mbf{e}_\mathrm{IMU}^{\mbf{C}}}{\partial\mbf{w}^\mbf{u}_\kmin}\right)^\trans.
\end{align}
The error Jacobian is derived to be
\begin{align}
    \f{\partial\mbf{e}_\mathrm{IMU}^{\mbf{C}}}{\partial\mbf{w}_\kmin^\mbf{u}} = - \mbf{J}_\ell^{-1} \left(\mbf{e}_\mathrm{IMU}^{\mbf{C}}\right)\mbf{J}_\ell(\mbf{u}_\kmin T_\kmin)T_\kmin,
\end{align}
where $\mbf{J}_\ell$ is the left group Jacobian of $SO(3)$. The position component error is computed as
\begin{align}
    \mbf{e}_\mathrm{IMU}^{\mbf{r}} = \mbf{v}_\kmin - \mbf{C}_\kmin^\trans\left(\f{\mbf{r}_k - \mbf{r}_\kmin}{T_\kmin}\right),
\end{align}
with a weighting resulting from $\mbf{v}_\kmin$ being $\mbs{\Sigma}_\mathrm{IMU}^\mbf{r} = \mbs{\Sigma}_\kmin^\mbf{v}$. The final IMU tracking loss is then
\begin{align}
    \mathcal{L}_\mathrm{IMU}^{\mathrm{track}} = \mbf{e}_\mathrm{IMU}^{{\mbf{C}}^\trans}\mbs{\Sigma}_\mathrm{IMU}^{\mbf{C}^{-1}}\mbf{e}_\mathrm{IMU}^{\mbf{C}} + \mbf{e}_\mathrm{IMU}^{{\mbf{r}}^\trans}\mbs{\Sigma}_\mathrm{IMU}^{\mbf{r}^{-1}}\mbf{e}_\mathrm{IMU}^{\mbf{r}}.
\end{align}

\subsubsection{Mapping with Motion Data}
In order to make use of IMU data in the mapping optimization, IMU preintegration is used. Introduced in \cite{Lupton2012}, IMU preintegration allows to group multiple IMU measurements into a relative motion increment (RMI). The RMI then functions as a single ``measurement'', defining a probabilistic motion constraint between two poses separated by any number of IMU measurements. Since the keyframes used for each mapping step are not predetermined, a potentially different number of IMU measurements can be connected to each consecutive keyframe. IMU preintegration greatly simplifies the computation of corresponding motion constraints between these keyframes and removes the need to keep track of every IMU measurement received. Since the same information is used, there is no change in performance.

The RMIs are computed incrementally between keyframes and are saved whenever a new keyframe is added. Consider the current camera pose $\mbf{X}_k$ at time $t_{k}$, the previous keyframe camera pose $\mbf{X}_i$ at time $t_i$, and the RMI $\Delta\mbf{X}_{ik}~=~(\Delta\mbf{C}_{ik}, \Delta\mbf{r}_{ik})$ connecting the two. The orientation component $\Delta\mbf{C}_{ik}$ and position component $\Delta\mbf{r}_{ik}$ are incrementally updated with the most recent IMU measurement that arrived at $t_\kmin \in [t_i, t_k]$ according to
\begin{align}
    \Delta\mbf{C}_{ik} &= \Delta\mbf{C}_{i\kmin}\Exp(\mbf{u}_\kmin T_\kmin),\\
    \Delta\mbf{r}_{ik} &= \Delta\mbf{r}_{i\kmin} + \Delta\mbf{C}_{i\kmin}(\mbf{v}_\kmin T_\kmin),
\end{align}
with full derivation omitted for brevity. Note, when a new keyframe is created, the RMI is reset to $\Delta\mbf{X}_{ik} = (\eye, \zero)$. The uncertainty on the RMI is also updated incrementally as
\begin{align}
    \mbs{\Sigma}_{ik}^\mathrm{RMI} &= \left(\f{\partial \Delta\mbf{X}_{ik}}{\partial \Delta\mbf{X}_{i\kmin}}\right)\mbs{\Sigma}_{i\kmin}^\mathrm{RMI}\left(\f{\partial \Delta\mbf{X}_{ik}}{\partial \Delta\mbf{X}_{i\kmin}}\right)^\trans \nonumber\\ &\qquad\qquad + \left(\f{\partial \Delta\mbf{X}_{ik}}{\partial \mbf{w}_\kmin}\right)\mbs{\Sigma}_{\kmin}^\mbf{w}\left(\f{\partial \Delta\mbf{X}_{ik}}{\partial \mbf{w}_\kmin}\right)^\trans,
\end{align}
where $\mbf{w}_\kmin = \begin{bmatrix}{\mbf{w}_\kmin^\mbf{u}}^\trans & {\mbf{w}_\kmin^\mbf{v}}^\trans \end{bmatrix}^\trans$ enters $\Delta\mbf{X}_{ik}$ through the measurements, and $\mbs{\Sigma}_{\kmin}^\mbf{w} = \diag(\mbs{\Sigma}_{\kmin}^\mbf{u}, \mbs{\Sigma}_{\kmin}^\mbf{v})$. As discussed in Section \ref{sec:impl_details}, $\mbs{\Sigma}_{ik}^\mathrm{RMI}$ was not used to downweight the mapping IMU loss, and thus its explicit form is omitted for brevity.

The final computed RMI is defined between two keyframes. In the case that, during optimization, involved keyframes are temporally separated by one or more unused keyframes, the total RMI can be computed by multiplying the individual RMIs together. For example, if keyframes 2 and 4 are involved in the optimization but 3 is not, the RMI between the involved keyframes can be computed as $\mathbf{X}_{24} = \mathbf{X}_{23}\mathbf{X}_{34}$. The uncertainty on the RMIs also needs to be propagated accordingly.

The RMI error between keyframe $i$ and $j$ is computed as
\begin{align}
    \mbf{e}_{ij}^{\Delta\mbf{C}} &= \Log\left(\Delta\mbf{C}_{ij}\mbf{C}_j^\trans\mbf{C}_i\right),\\
    \mbf{e}_{ij}^{\Delta\mbf{r}} &= \Delta\mbf{v}_{ij} - \mbf{C}_i^\trans\left(\mbf{r}_j - \mbf{r}_i\right),\\
    \mbf{e}_{ij}^\mathrm{RMI} &= \begin{bmatrix} \mbf{e}_{ij}^{\Delta\mbf{C}^\trans} & \mbf{e}_{ij}^{\Delta\mbf{r}^\trans}\end{bmatrix}^\trans,
\end{align}
with a corresponding weight $\mbs{\Sigma}_{ij}^{\mathrm{RMI}}$. The total final RMI loss for some subset of keyframes $z$, for example $z \in [2, 4]$ in the example above, is written as
\begin{align}\label{eq:RMI_map_loss}
    \mathcal{L}_\mathrm{RMI}^{\mathrm{map}} = \sum_{q = 2}^{\mathrm{len}(z)} \mbf{e}_{z[q-1]z[q]}^{\mathrm{RMI}^\trans} \mbs{\Sigma}_{z[q-1]z[q]}^{\mathrm{RMI}^{-1}} \mbf{e}_{z[q-1]z[q]}^{\mathrm{RMI}}.
\end{align}

\subsection{Background Model}\label{sec:bgModel}

For SLAM-based algorithms, encoding distant features into the map can lead to significant improvements in orientation estimation accuracy \cite{near_dist_bearing_slam}. Conversely, the original NICE-SLAM algorithm can only encode map features that lie within a (restrictive) bounding box. In order to extend NICE-SLAM to \textit{unbounded} environments without requiring unbounded memory resources, the background of a given scene is encoded using a spherical grid. This representation is similar to the Multi-Sphere Images (MSI) proposed in \cite{kaizhang2020} and used in \cite{yu_plenoxels_2021}. 

\subsubsection{Modeling a Sphere at Infinity}\label{sec:infSphr}
In order to preserve real-time capability by keeping the representation computationally light, the background is chosen to be modelled with only one background sphere. This sphere is modeled as if it is infinitely far away from the central world frame and, by extension, any given camera frame. This allows distant objects to be represented on the horizon, which can be very helpful for localizing the camera's orientation. Modelling the background as a sphere simplifies computations, since only the direction of the pixel ray is required to evaluate the contribution of the background model. To see why this is true, consider a point on the background sphere along a given pixel ray in the camera frame, $ \mbf{x}_c = \beta \hat{\mbf{x}}_c$, where $\hat{\mbf{x}}_c$ is the normalized direction of the ray and $\beta$ represents the magnitude of the ray. The normalized ray in the world frame can be expressed as
\begin{equation}
    \hat{\mbf{x}}_w  = \f{\mbf{x}_w}{\norm{\mbf{x}_w}_2} =\f{\mbf{C}_{wc} \mbf{x}_c + \mbf{r}^{cw}_w}{\norm{\mbf{x}_c + \mbf{r}^{cw}_w}_2} = \f{ \mbf{C}_{wc} \beta \hat{\mbf{x}}_c + \mbf{r}^{cw}_w}{\beta\norm{\hat{\mbf{x}}_c + \mbf{r}^{cw}_w/\beta}_2 }.
\end{equation}
Now, as $ \beta \rightarrow \infty $, $\hat{\mbf{x}}_w \simeq \mbf{C}_{wc} \hat{\mbf{x}}_c $, that is, the sphere only needs to be evaluated along the rotated camera frame ray.
\begin{figure*}[ht!]
\centering
\begin{subfigure}{0.43\linewidth}
    \includegraphics[width=\linewidth, trim={8cm 4cm 8cm 5cm},clip]{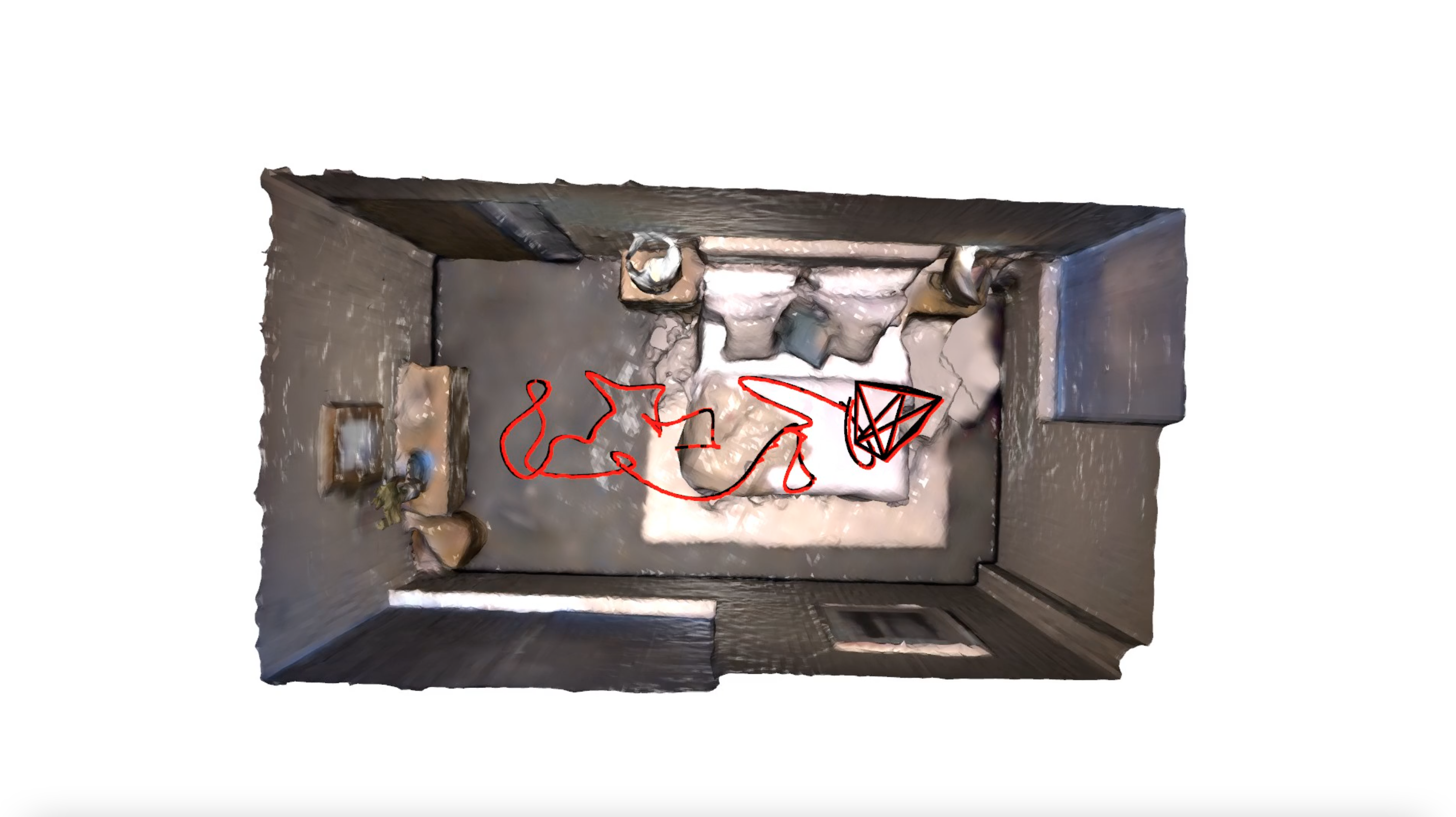}
    \caption{\textbf{NICE-SLAM:} Full number of iter.}
\end{subfigure}
\begin{subfigure}{0.43\linewidth}
    \includegraphics[width=\linewidth, trim={8cm 4cm 8cm 5cm},clip]{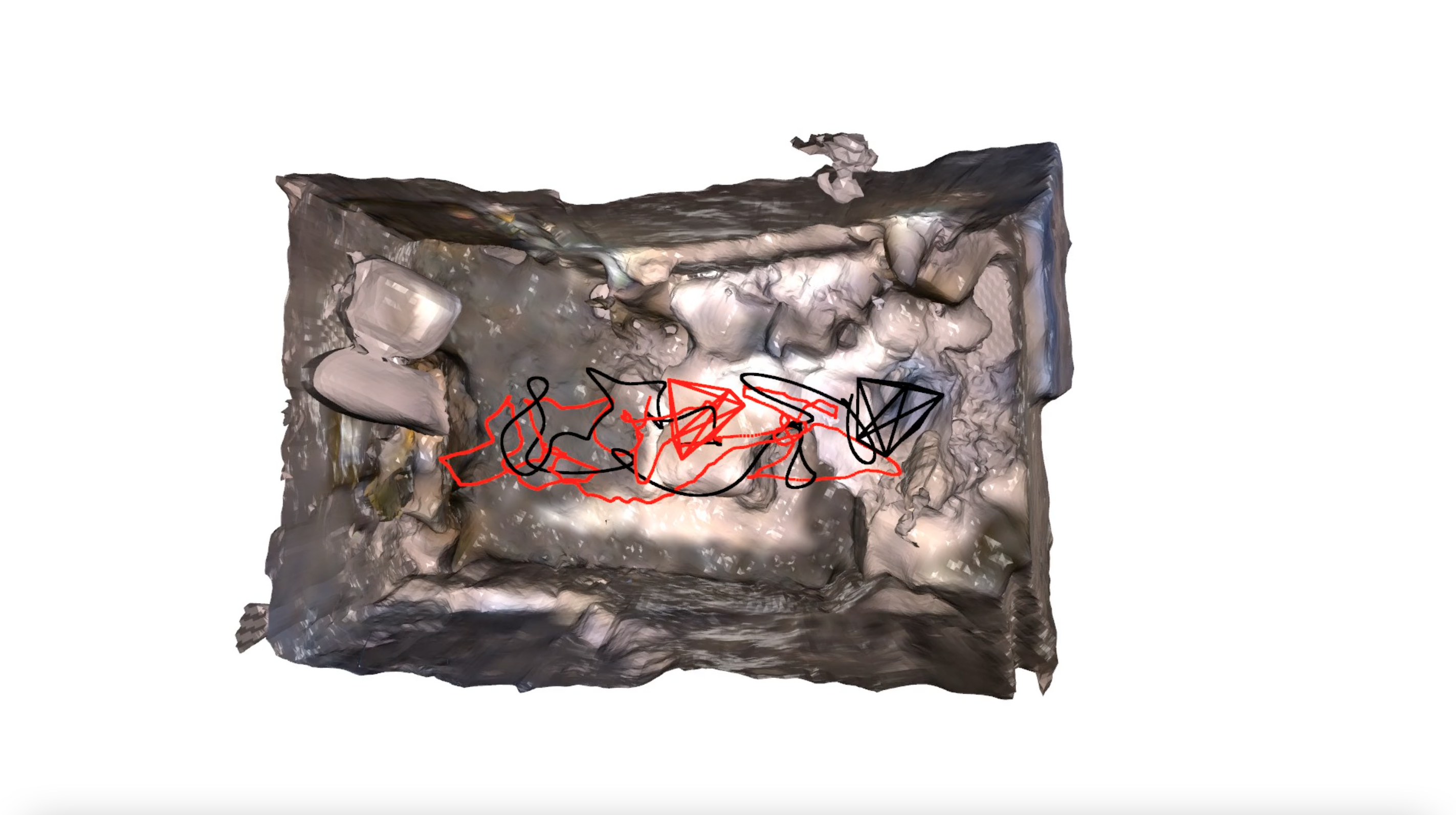}
    \caption{\textbf{NICE-SLAM:} Low number of iter.}
\end{subfigure}
\begin{subfigure}{0.43\linewidth}
    \includegraphics[width=\linewidth, trim={8cm 4cm 8cm 5cm},clip]{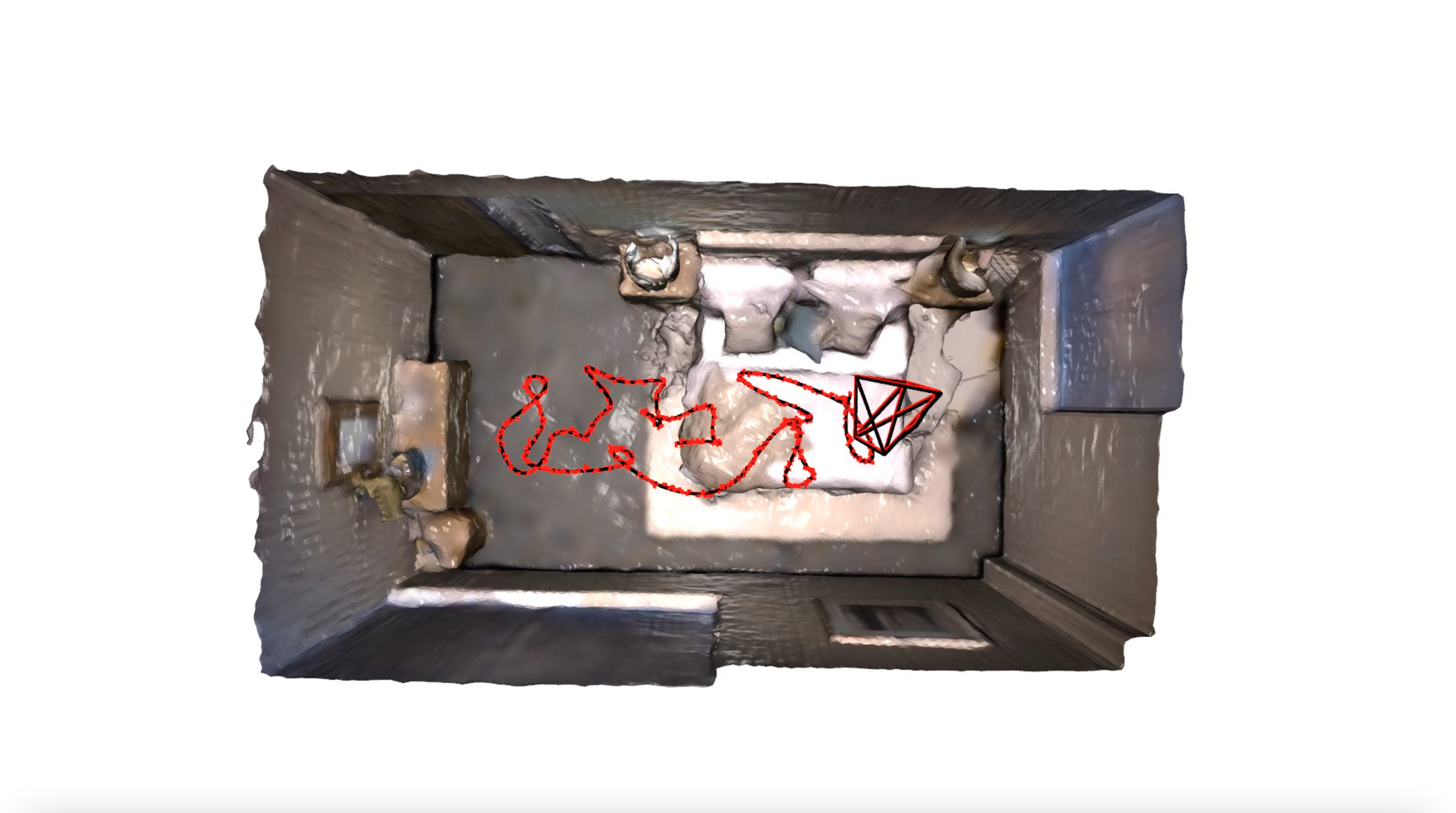}
    \caption{\textbf{Ours:} Full number of iter. (IMU + Dep.).}
\end{subfigure}
\begin{subfigure}{0.43\linewidth}
    \includegraphics[width=\linewidth, trim={8cm 4cm 8cm 5cm},clip]{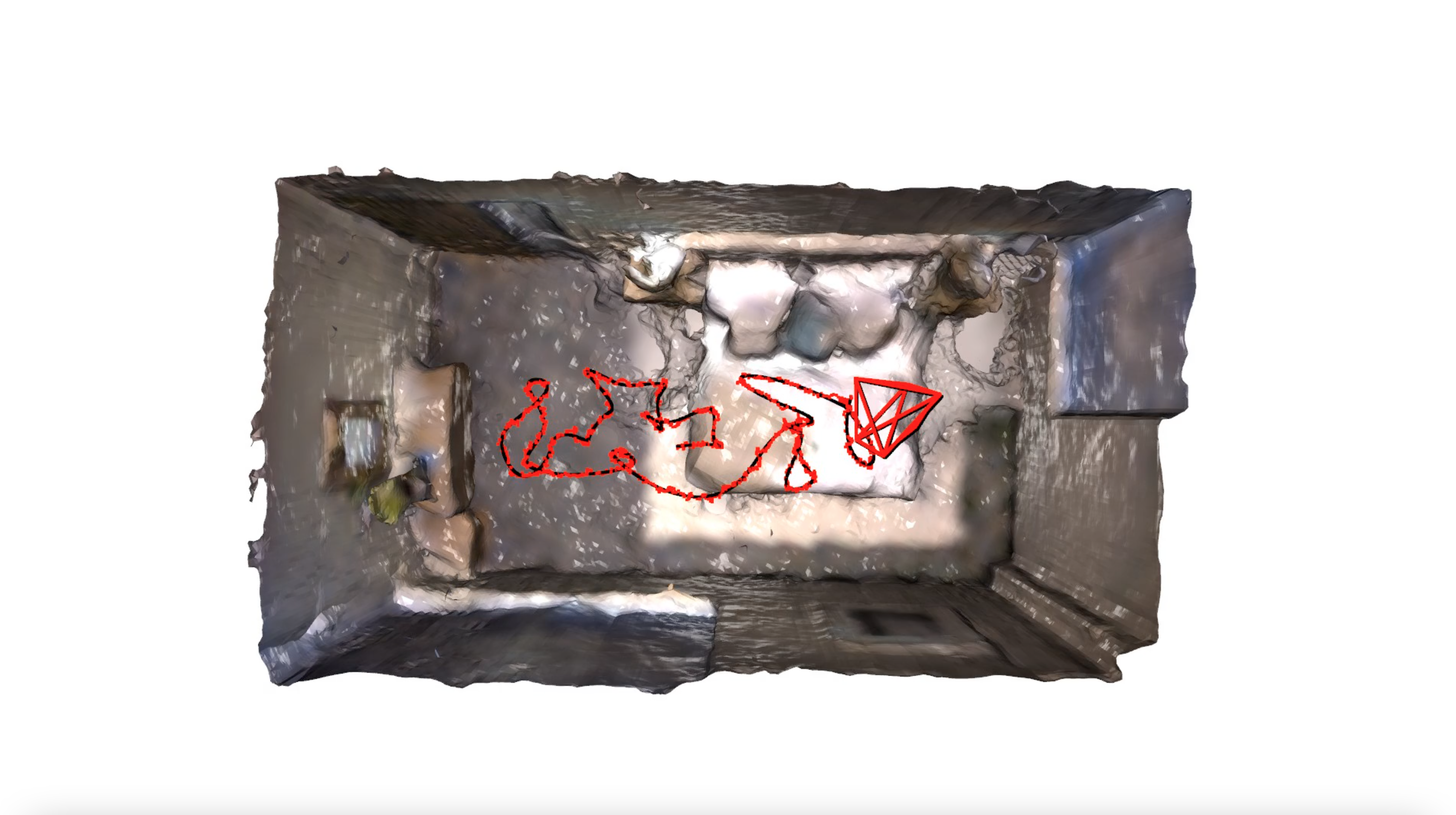}
    \caption{\textbf{Ours:} Low number of iter. (IMU + Dep.).}
\end{subfigure}
\caption{Reconstruction results for default and modified NICE-SLAM run at a full (10/60) and low (5/10) number of iterations for tracking/mapping on the ``room1'' Replica scene. The black and red trajectories are the ground truth and estimate respectively.}
\label{fig:room1}
\vspace{-4mm}
\end{figure*}
\subsubsection{Background NeRF Model}
The pixel colour from a given ray, $ \mbf{x}_w(\beta) = \mbf{r}^{cw}_w + \beta \hat{\mbf{x}}_w $\footnote{This is exactly the ray $\mbf{r}(t) = \mbf{o} + t\mbf{d}$ from the original NeRF paper \cite{mildenhall_nerf_2020}.}, is given by
\begin{gather}
    \hat{I}_n = \int_{\beta_n}^{\beta_f} T(\beta)\sigma(\mbf{x}_w(\beta))\mbf{c}(\mbf{x}_w(\beta)) d\beta, \\
    \mbox{where} \quad\quad T(\beta) = \Exp\left(-\int_{\beta_n}^{\beta_f}\sigma(\mbf{x}_w(s))ds\right)
\end{gather}
represents the accumulated transmittance (probability that a ray travels from $ \beta $ to $\beta_n$), $ \sigma(\mbf{x}) $ represents the scene volume density, $\mbf{c}(\mbf{x}_w(\beta))$ represents the colour at a given point and $ \beta_n $, $ \beta_f $, represent the near and far limits of the depth parameter $\beta$, respectively. Now, suppose the far limit extends to infinity, $ \beta_f \rightarrow \infty$. The integral then becomes
\begin{align}
    \hat{I}_{n,\infty} &= \int_{\beta_n}^{\infty} T(\beta)\sigma(\mbf{x}_w(\beta))\mbf{c}(\mbf{x}_w(\beta)) d\beta \\ 
    &= \int_{\beta_n}^{\beta_f} T(\beta)\sigma(\mbf{x}_w(\beta))\mbf{c}(\mbf{x}_w(\beta)) d\beta \nonumber\\
    &\qquad\qquad +\int_{\beta_f}^{\infty} T(\beta)\sigma(\mbf{x}_w(\beta))\mbf{c}(\mbf{x}_w(\beta)) d\beta \\ 
    &= \hat{I}_n + T(\beta_f) \mbf{c}_{\infty}(\mbf{x}_w(\beta)).
\end{align}
where $\hat{I}_n$ is the original NICE-SLAM \textit{foreground} NeRF and $\mbf{c}_{\infty}(\mbf{x}_w(\beta)) = \int_{\beta_f}^{\infty} T(\beta)\sigma(\mbf{x}_w(\beta))\mbf{c}(\mbf{x}_w(\beta)) d\beta$ represents the evaluation from $\beta_f$ to $\infty$ of a separate \textit{background} NeRF. This makes use of the fact that
\begin{align*}
    \log&(T(c)) = -\int_{a}^{c}\sigma(\mbf{x}_w(s))ds\\ 
    &=\-\int_{a}^{b}\sigma(\mbf{x}_w(s))ds -\int_{b}^{c}\sigma(\mbf{x}_w(s))ds=\log(T(b)T(b,c)).
\end{align*}
Thus, the background of the scene can be represented by simply adding a defined \textit{background} colour, $\mbf{c}_{\infty}(\mbf{x}_w(\beta))$, multiplied by the final transmittance value at the boundary of the foreground NeRF, $T(\beta_f)$. It is assumed that the background is mostly empty except for some points that are very far away. As shown in Section \ref{sec:infSphr}, the background colour can be represented entirely in terms of the ray direction: $\mbf{c}_{\infty}(\mbf{x}_w(\beta)) \simeq \mbf{c}_{\infty}(\hat{\mbf{x}}_w) = \mbf{c}_{bg}(\hat{\mbf{x}}_w)$. 

Similar to the encoded color grid for NeRF-SLAM \cite{rosinol_nerf-slam_2022}, a 2D grid of background features that are warped onto a sphere is generated. At render time, the grid and a pixel ray, $\hat{\mbf{x}}_w$, are used to interpolate a feature. Then, the feature is decoded and scaled by the boundary transmittance of the foreground NeRF, $T(\beta_f)$. Note that NICE-SLAM already computes the world frame rays, $ \hat{\mbf{x}}_w $, and boundary transmittance, $T(\beta_f)$, so no additional cost is incurred to use them. The specific implementation of the background sphere in this paper is given in Section \ref{sec:bgImp}, although many other parametrizations are also possible.

\renewcommand{\arraystretch}{1.25}
\begin{table*}[ht!]
\centering
\begin{tabular}{||c||c||c|c|c|c|c|c||c|c|c|c|c|c||}
\cline{3-14}
\cline{3-14}
\multicolumn{1}{c}{} & \multirow{4}{*}{} & \multicolumn{6}{c||}{Room 0} & \multicolumn{6}{c||}{Room 1} \\
\cline{3-14}
\multicolumn{1}{c}{} & & \multicolumn{2}{c|}{Tracking} & \multicolumn{4}{c||}{Mapping} & \multicolumn{2}{c|}{Tracking} & \multicolumn{4}{c||}{Mapping} \\
\cline{3-14}
\multicolumn{1}{c}{} & & RMSE & Max. & Acc. & Comp. & C.R. & Dep. L1 & RMSE & Max. & Acc. & Comp. & C.R. & Dep. L1\\
\multicolumn{1}{c}{} & & [\si{cm}] $\downarrow$ & [\si{cm}] $\downarrow$ & [\si{cm}] $\downarrow$ & [\si{cm}] $\downarrow$ & [\%] $\uparrow$ & [\si{cm}] $\downarrow$ & [\si{cm}] $\downarrow$ & [\si{cm}] $\downarrow$ & [\si{cm}] $\downarrow$ & [\si{cm}] $\downarrow$ & [\%] $\uparrow$ & [\si{cm}] $\downarrow$\\
\hline
\hline
\multirow{4}{*}{\rotatebox[origin=c]{90}{Full Iter.}} & NICE-SLAM & 0.020 & 0.072 & 2.744 & 3.000 & 91.03 & 1.922 & 0.026 & 0.103 & 2.796 & 2.376 & 92.43 & 1.732\\
\cline{2-14}
& Depth & 0.018 & 0.074 & {\color{red}\textbf{2.646}} & {\color{blue}\textbf{2.840}} & {\color{blue}\textbf{91.25}} & {\color{blue}\textbf{1.763}} & 0.018 & 0.048 & {\color{red}\textbf{2.430}} & 2.257 & {\color{red}\textbf{93.54}} & {\color{red}\textbf{1.488}}\\
\cline{2-14}
 & IMU & {\color{blue}\textbf{0.003}} & {\color{red}\textbf{0.014}} & {\color{blue}\textbf{2.645}} & {\color{red}\textbf{2.860}} & {\color{red}\textbf{91.20}} & {\color{red}\textbf{1.773}} & {\color{red}\textbf{0.003}} & {\color{blue}\textbf{0.020}} & 2.505 & {\color{blue}\textbf{2.229}} & {\color{blue}\textbf{93.77}} & {\color{blue}\textbf{1.440}}\\
\cline{2-14}
& IMU + Dep. & {\color{red}\textbf{0.003}} & {\color{blue}\textbf{0.014}} & 2.651 & 2.977 & 90.78 & 2.010 & {\color{blue}\textbf{0.003}} & {\color{red}\textbf{0.026}} & {\color{blue}\textbf{2.308}} & {\color{red}\textbf{2.240}} & 93.10 & 1.493\\
\hline
\hline
\multirow{4}{*}{\rotatebox[origin=c]{90}{Low Iter.}} & NICE-SLAM & 0.156 & 0.627 & 5.083 & 5.030 & 67.29 & 5.677 & 0.531 & 1.148 & 15.55 & 16.67 & 35.68 & 24.98\\
\cline{2-14}
& Depth & 0.211 & 0.539 & 5.641 & 6.330 & 66.13 & 7.937 & 0.343 & 0.820 & 18.17 & 18.57 & 28.79 & 25.89\\
\cline{2-14}
& IMU & {\color{red}\textbf{0.002}} & {\color{blue}\textbf{0.006}} & {\color{red}\textbf{2.972}} & {\color{red}\textbf{3.272}} & {\color{red}\textbf{88.28}} & {\color{blue}\textbf{2.420}} & {\color{blue}\textbf{0.002}} & {\color{blue}\textbf{0.021}} & {\color{blue}\textbf{2.184}} & {\color{red}\textbf{2.605}} & {\color{red}\textbf{91.25}} & {\color{blue}\textbf{1.824}}\\
\cline{2-14}
& IMU + Dep. & {\color{blue}\textbf{0.002}} & {\color{red}\textbf{0.006}} & {\color{blue}\textbf{2.918}} & {\color{blue}\textbf{3.120}} & {\color{blue}\textbf{88.84}} & {\color{red}\textbf{2.450}} & {\color{red}\textbf{0.002}} & {\color{red}\textbf{0.022}} & {\color{red}\textbf{2.239}} & {\color{blue}\textbf{2.501}} & {\color{blue}\textbf{91.94}} & {\color{red}\textbf{1.922}}\\
\hline
\hline
\multicolumn{2}{||c||}{IMU Dead Reck.} & 0.027 & 0.072 & $\sim$ & $\sim$ & $\sim$ & $\sim$ & 0.034 & 0.095 & $\sim$ & $\sim$ & $\sim$ & $\sim$\\
\hline
\hline
\end{tabular}
\caption{Results from two trials on the ``Replica'' dataset. Best pre-rounded results within each quadrant are {\color{blue}\textbf{blue}}, with second best in {\color{red}\textbf{red}}. \textbf{RMSE} is the Root Mean Squared Error; \textbf{Max.} is the worst case tracking error; \textbf{Accuracy (Acc.)} is the average distance to the nearest ground truth mesh from the reconstructed mesh; \textbf{Completion (Comp.)} is the average distance to the nearest reconstructed mesh from the ground truth mesh; \textbf{C.R.} is the completion ratio, or percentage of points with completion under 5 cm; \textbf{Dep. L1} is the sum of absolute depth errors.}
\label{tbl:resultsREP}
\vspace{-5mm}
\end{table*}

\vspace{-2mm}
\section{Results}
The changes outlined in this paper are implemented using the existing NICE-SLAM codebase \cite{zhu_nice-slam_2022}, with modifications made to the rendering and loss functions as appropriate. 
\subsection{Dataset Selection}
Three datasets are selected to verify the performance of the enhancements detailed in this paper. The first two datasets, ``room0'' and ``room1'' from the Replica dataset \cite{replica_dataset}, are selected for ease of comparison with previous studies \cite{zhu_nice-slam_2022,sucar_imap_2021}. Due to the limited size of the rooms in the Replica dataset, it was not practical to split the scene into an explicit background and foreground. Thus, to showcase the benefit of including the background sphere, the ``freiburg2\_rpy'' scene from the TUM-RGBD dataset \cite{sturm12iros} is included as it has a larger geometry.

While it is often possible to capture a given scene with a single foreground NeRF, this approach is not always feasible for larger or unbounded scenes due to memory constraints \cite{kaizhang2020}. The loss induced by limiting the foreground model can be mitigated by including a background model. The ``freiburg2\_rpy'' dataset consists of a desk in the foreground, with the wall of the room in the distant background. For this dataset, the foreground NeRF is limited to a 6.0 m cube around the starting position in order to encompass just foreground objects. This allows to simulate a larger or unbounded scene and evaluate any performance gain when the background model is included in NICE-SLAM. 

The TUM-RGBD dataset does not include ground truth meshes, so the constructed map is evaluated only in terms of the average colour and depth loss obtained between the recorded images and the reconstructed map.

\subsection{Implementation Details}\label{sec:impl_details}
To test the impact of the listed changes on robustness, the algorithm is tested with the full, default number of iterations suggested by the authors of NICE-SLAM and with a heavily reduced number of iterations to approach real-time performance. Specifically, for the Replica datasets, the ``full number of iterations'' corresponds to 10 and 60 iterations for tracking and mapping respectively, whereas the ``low number of iterations'' corresponds to 5 and 10 iterations. This change was empirically observed to reduce the overall runtime of the considered Replica scenes from approximately 48 minutes to 10 minutes on the NVIDIA RTX 2070 SUPER GPU. All other parameters are held constant across different trials.

In practice, an RMI connection to the most recently tracked state is not included in the mapping loss due to the parallel nature of the tracking and mapping threads. Additionally, although the RMI mapping loss weight $\mbs{\Sigma}_{ik}^\mathrm{RMI}$ was computed, it was found that manual tuning of the loss weighting was still required. This is likely because other parts of the mapping loss are weighted via hand-tuned parameters. If it were possible to compute the uncertainty for all loss components, then $\mbs{\Sigma}_{ik}^\mathrm{RMI}$ could become useful. It was found that simply omitting $\mbs{\Sigma}_{ik}^\mathrm{RMI}$ from the loss in \eqref{eq:RMI_map_loss} yielded consistent results.

\subsubsection{Depth Sensor Modelling}
The considered datasets do not include information about the noise parameters of the specific sensors that were used. As a result, the depth uncertainty used in the experiments is approximated based on any available online technical specifications of the sensor. This information is limited, and highlights the need for future datasets to explicitly include sensor noise properties. 

\subsubsection{IMU Modelling}\label{sec:imu_modelling}
Since IMU data is not considered in NICE-SLAM, it is not included with any of the provided datasets. Therefore, data is generated at 100 Hz from the provided ground truth poses for all considered datasets and corrupted with white, Gaussian noise. In this paper, ``IMU'' measurements refer to linear and angular velocity measurements. This is done in order to simplify derivations, since camera velocity estimates are not included, and to avoid doubly differentiating and then integrating acceleration data, which results in very noisy measurements.

A linear velocity measurement $\mbf{v}_k$ at time $t_k$ is generated from ground truth camera poses according to
\begin{align}
    \mbf{v}_k = \mbfbar{v}_k + \mbf{w}^\mbf{v}_k= \mbf{C}_{k}^\trans \left(\f{\mbf{r}_{k+1} - \mbf{r}_k}{T_k}\right) + \mbf{w}^\mbf{v}_k,\label{eq:lin_vel_eq}
\end{align}
where $\mbfbar{v}_k$ is the true linear velocity, $\mbf{C}_{k} \in SO(3)$ is the camera orientation at $t_k$, $\mbf{r}_\imath \in \mathbb{R}^3$ is the camera position at $t_\imath, \imath \in [k, k+1]$, $T_k = t_{k+1} - t_k$ is the time increment, and $\mbf{w}^\mbf{v}_k \sim \mathcal{N}(\mbf{0}, \mbs{\Sigma}_k^\mbf{v})$ is the measurement noise with covariance $\mbs{\Sigma}_k^\mbf{v}$. An angular velocity measurement $\mbf{u}_k$ at time $t_k$ is generated from ground truth camera poses according to
\begin{align}
    \mbf{u}_k = \mbfbar{u}_k + \mbf{w}_k^\mbf{u}= \f{\Log(\mbf{C}_k^\trans\mbf{C}_{k+1})}{T_k} + \mbf{w}_k^\mbf{u},\label{eq:ang_vel_eq}
\end{align}
where $\mbfbar{u}_k$ is the true angular velocity and $\mbf{w}^\mbf{u}_k \sim \mathcal{N}(\mbf{0}, \Sigma_k^\mbf{u})$ is the measurement noise with covariance $\Sigma_k^\mbf{u}$. These models assume that the velocity measurements are constant over $T_k$.

\subsubsection{Background Sphere Modelling}\label{sec:bgImp}

In our implementation, the background sphere is modelled using unit spherical coordinates, parameterized in terms of a 2-dimensional grid. Given a sampled unit ray direction, $ \hat{\mbf{x}}_w \in \mathbb{R}^3$, the unit ray is first converted to polar angle ($\theta$) and azimuth ($\phi$) parameters using the following transformation:
\begin{gather}
    \theta = \cos^{-1}(\hat{\mbf{x}}_{w,3}), \\
    \phi = \mbox{sign}(\hat{\mbf{x}}_{w,2}) \cos^{-1}(\hat{\mbf{x}}_{w,1}).
\end{gather}
The background NeRF is then evaluated using a 2-dimensional version of the interpolation scheme used in \cite{zhu_nice-slam_2022}. 
The background NeRF decoder parameters are set to the same as those of the colour NeRF.


\subsection{Depth Uncertainty and IMU Results}
A full ablation study for depth uncertainty and IMU data is shown in Table \ref{tbl:resultsREP} and Figure \ref{fig:room1}.  It can be seen in Table \ref{tbl:resultsREP} that accounting for depth uncertainty moderately improves most metrics in both evaluation scenes. The effect of including depth uncertainty is less clear when IMU data is used. This is likely due to the relatively small scale of the scenes and the approximations made when modelling the depth sensor due to a lack of sensor noise information in the datasets. 

Inclusion of motion data leads to drastic improvement in tracking error for all cases and the map reconstruction error with a reduced number of iterations. For the ``room0'' scene, the tracking error drops by 85\% and 97\% with the default and limited number of iterations respectively. This effect is even larger in the ``room1'' scene, which has a more challenging motion profile. The impact of motion data in situations with limited resources is visualized in Figure \ref{fig:room1}, where the default NICE-SLAM algorithm fails to construct a usable map whereas the proposed modifications yield a map of nearly identical quality. It is therefore concluded that the addition of IMU data can lead to significantly faster convergence of NICE-SLAM.
Trajectory results from IMU-only dead reckoning are also included in Table \ref{tbl:resultsREP} to demonstrate that, by itself, the simulated IMU data is not sufficient to achieve good tracking.

The combined effect of depth uncertainty and IMU measurements is also considered in the background sphere experiments as shown in Table \ref{tbl:resultsTUM} and Figure \ref{fig:bg_grid_compare}. A similar effect can be observed: the inclusion of the two changes drastically improves the tracking performance and also lowers the colour and depth cost of the reconstructed map.

\subsection{Background Sphere Results}

\begin{figure*}[ht!]
\centering
\includegraphics[width=0.87\textwidth]{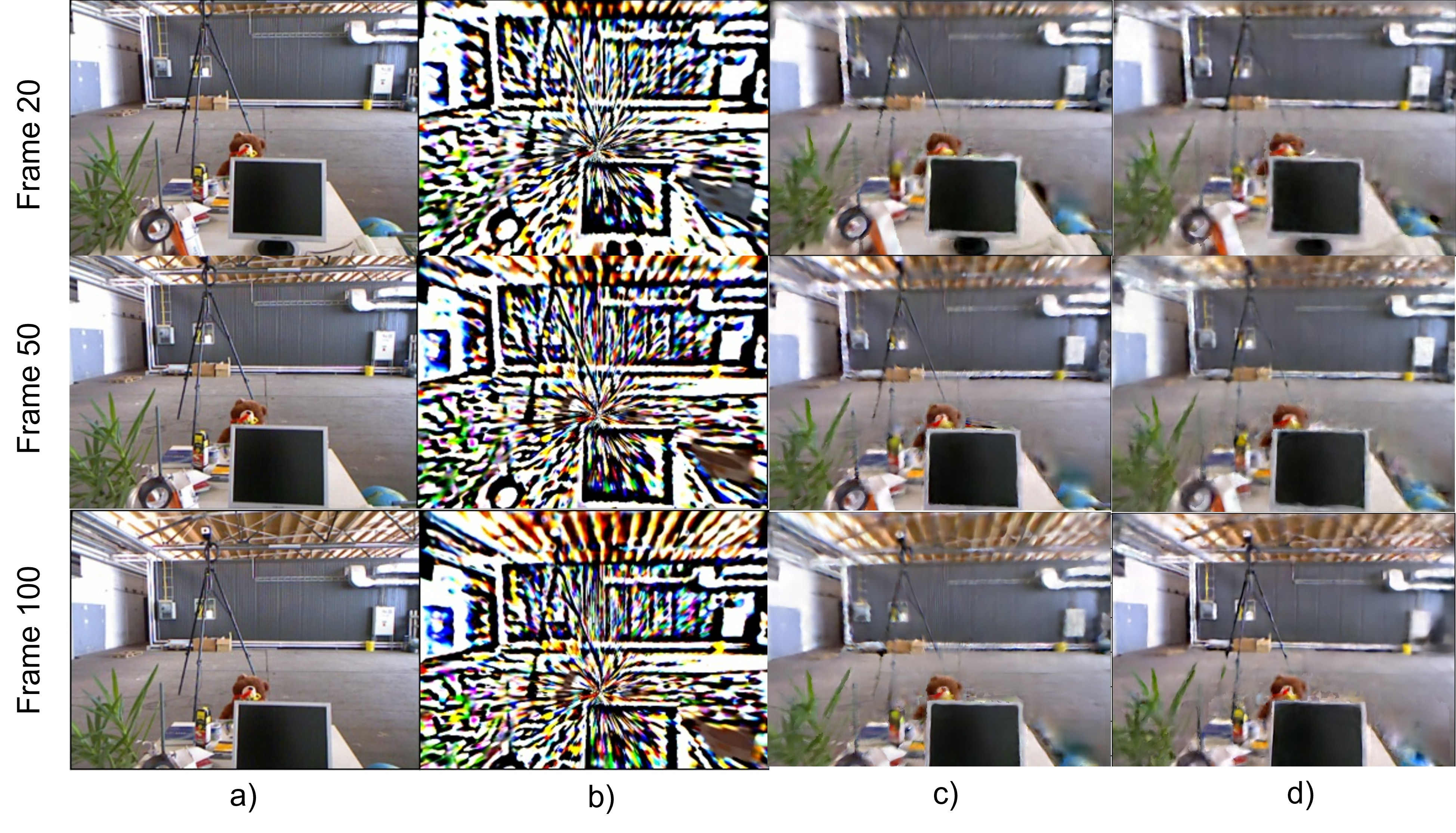}
\vspace*{-10pt}
\caption{The effect of the background sphere across different frames in the ``freiburg2\_rpy'' scene. \textbf{a)} Input image; \textbf{b)} Background sphere only; \textbf{c)} NICE-SLAM with background active; \textbf{d)} Default NICE-SLAM (no background model).}
\label{fig:bg_grid_compare}
\vspace*{-20pt}
\end{figure*}

Figure \ref{fig:bg_grid_compare} demonstrates the background model and image reconstruction with and without a background sphere across three scenes in the dataset. By comparing columns c) and d) with a), it can be seen that the background model allows the details in the background to be represented more accurately in the earlier frames. In later frames, it seems that NICE-SLAM can approximate the background details equally well. This may be because there is not much translational motion in this dataset, meaning that the background can be adequately represented by the foreground. A lack of datasets with large scenes suitable for NeRF-SLAM evaluation is needed for further evaluation.

An ablation study that considers the effect of the background sphere is given in Table \ref{tbl:resultsTUM}. The RMSE and maximum tracking errors are considerably reduced when the background sphere is used. It is further reduced when IMU data is additionally included. As expected, the colour loss decreases when the background model is introduced in otherwise identical loss functions, reinforcing the proposition that the background model can improve overall scene representation. Note, an increase in the colour and depth cost when IMU data is included is expected as the overall objective function aims to minimize an additional loss.

\subsection{Combined Results}
Table \ref{tbl:resultsTUM} also presents results when all of the features described in this paper are used together. The RMSE clearly decreases in comparison to the default version of NICE-SLAM, indicating that the enhancements do improve overall performance. However, the inclusion of depth uncertainty in the cost decreases both tracking and mapping performance. This is hypothesized to either be a result of poor uncertainty modelling on account of a lack of sensor data, or due to some unforeseen interaction between the depth and the background sphere. Unfortunately, due to a lack of better scenes with a clear foreground and background, exploration of the underlying cause for the slight drop in performance from combining the features is left as future work. 
\renewcommand{\arraystretch}{1.25}
\begin{table}[t!]
\centering
\begin{tabular}{||c||c|c|c|c||}
\cline{2-5}
\cline{2-5}
\multicolumn{1}{c||}{} & \multicolumn{2}{c|}{Tracking} & \multicolumn{2}{c||}{Mapping}\\
\cline{2-5}
\multicolumn{1}{c||}{}& RMSE & Max. & Col. L1 & Dep. L1 \\
\multicolumn{1}{c||}{} & [\si{cm}] $\downarrow$ & [\si{cm}] $\downarrow$ & $\downarrow$ & $\downarrow$\\
\hline
\hline
Fg. (NICE-SLAM) & 0.016 & 0.038 & 145.1 & 4.117\\
\hline
Fg.+ IMU & {\color{blue}\textbf{0.006}} & {\color{blue}\textbf{0.019}} & 143.7 & 4.156 \\
\hline
Fg. + IMU + Dep. & 0.009 & 0.042 & 180.3 & 4.830 \\
\hline
Fg. + Bg. & 0.008 & 0.033 & {\color{blue}\textbf{140.5}} & {\color{blue}\textbf{3.322}}\\
\hline
Fg. + Bg. + IMU & {\color{red}\textbf{0.006}} & {\color{red}\textbf{0.027}} & {\color{red}\textbf{141.4}} & {\color{red}\textbf{3.588}}\\
\hline
Fg. + Bg. + IMU + Dep. & 0.011 & 0.050 & 177.0 & 4.176\\
\hline
\hline
\end{tabular}
\caption{Results from the ``freiburg2\_rpy'' scene. Best pre-rounded results are {\color{blue}\textbf{blue}}, with second best in {\color{red}\textbf{red}}. \textbf{RMSE} is the Root Mean Squared Error; \textbf{Max.} is the worst case tracking error; \textbf{Col. L1} is the avg. colour loss; \textbf{Dep. L1} is the avg. depth loss. ``Fg.'' and ``Bg.'' refer to the use of foreground and spherical background NeRF models, respectively. The original NICE-SLAM is equivalent to ``Fg.'' only.}
\label{tbl:resultsTUM}
\vspace{-6mm}
\end{table}

\vspace{-1.5mm}

\section{Conclusion}
It is shown that the addition of depth uncertainty and motion data can improve the accuracy of a NeRF-based SLAM algorithm, especially when real-time performance is desired. It is also demonstrated that a spherical background model can improve the image reconstruction of a NeRF map when the scene is too large to be modeled by a bounded foreground NeRF. In order for NeRF-based SLAM to approach the usability of existing dense SLAM algorithms, it must become possible to construct NeRF map representations of truly unbounded scenes in real time. The improvements presented here begin to approach this goal. One future avenue of interest is to leverage ideas in \cite{vox_fusion} to avoid using a predefined grid. Additionally, this work found a need for additional datasets geared towards NeRF-based SLAM methods, specifically for those that provide outdoor mesh models, motion data, and fully characterized sensors.





\section*{Acknowledgements}
Thank you to Alec Krawciw for his help in facilitating our experiments.

\bibliographystyle{IEEEtran}
\bibliography{IEEEabrv,lib,DenseSLAM}

\end{document}